\documentclass[single-column]{fairmeta}
\usepackage[most]{tcolorbox}
\usepackage{amsmath}
\usepackage{amssymb}
\usepackage{listings}
\usepackage{multirow}
\usepackage{makecell}
\usepackage{graphicx}
\usepackage{wrapfig}
\usepackage{float}
\usepackage{tabularx}
\usepackage{textcomp}
\usepackage{stfloats}
\usepackage{url}
\usepackage{verbatim}
\usepackage{titlesec}
\usepackage{tocloft}
\usepackage{adjustbox}
\usepackage{pifont}
\usepackage{tikz}
\usepackage{comment}
\usepackage{colortbl}
\usepackage{color}
\usepackage{booktabs}
\usepackage{subcaption}
\usepackage[table]{xcolor}
\usepackage{xltabular}
\usepackage{array}
\usepackage{threeparttable}
\usepackage{diagbox}
\usepackage{pgfplots}
\pgfplotsset{compat=1.18}

\RequirePackage{xspace}
\makeatletter
\DeclareRobustCommand\onedot{\futurelet\@let@token\@onedot}
\def\@onedot{\ifx\@let@token.\else.\null\fi\xspace}
\makeatother

\definecolor{wmgreen}{RGB}{46,125,50}
\definecolor{wmorange}{RGB}{230,145,56}
\definecolor{wmgray}{RGB}{120,120,120}

\title{A Cookbook of 3D Vision: Data, Learning Paradigms, and Application}

\author[1,*]{Hongyang~Du}
\author[2,*]{Zongxia~Li}
\author[3,*]{Dawei~Liu}
\author[4,*]{Runhao~Li}
\author[5]{Haoyuan~Song}
\author[5]{Qingyu~Zhang}
\author[5]{Yubo~Wang}
\author[1]{Jingcheng~Ni}
\author[1]{Shihang~Gui}
\author[6]{Congchao~Dong}
\author[7]{Tao~Hu}

\small{
\affiliation[1]{Brown~University}
\affiliation[2]{University~of~Maryland,~College~Park}
\affiliation[3]{University~of~Pennsylvania}
\affiliation[4]{University~of~Southern~California}
\affiliation[5]{New~York~University}
\affiliation[6]{The~University~of~Sydney}
\affiliation[7]{Stability~AI}
}

\footnotesize{\contribution[*]{Equal Contribution}}

\abstract{%
3D vision has rapidly evolved, driven by increasingly diverse data representations, learning paradigms, and modeling strategies.
Yet the field remains fragmented across representations and benchmarks, making it difficult to develop unified perspectives on efficiency, fidelity, and scalability.
This work provides a data-centric taxonomy of 3D vision that connects geometric representations, datasets, learning frameworks, and applications within a single conceptual map.
We begin by analysing the principal structural representations of 3D data---point clouds, meshes, voxels, and 3D Gaussians---along with their acquisition pipelines. We then examine how dataset design, benchmark construction, and supervision regimes shape recent advances, spanning 2D-supervised 3D learning, implicit neural representations, and 4D world modeling.
Through this integrative lens, we clarify the relationships among representations, learning paradigms, and downstream tasks in reconstruction, generation, and video modeling, offering a consolidated view of emerging trends toward balancing efficiency and fidelity and toward multimodal geometric grounding.

}

\correspondence{%
  Hongyang Du <\email{hongyang\_du@brown.edu}> and Zongxia Li <\email{zli12321@umd.edu}>  }

\metadata[Repository]{\url{https://github.com/Hongyang-Du/awesome-3d-datasets}}

\begin{document}

\maketitle

\section{Introduction}

3D vision has emerged as a central pillar in modern computer vision, with widespread applications in autonomous navigation~\cite{mur2017orbslam2}, robotic manipulation~\cite{mahler2017dexnet}, augmented reality~\cite{klein2007ptam,izadi2011kinectfusion}, and digital reconstruction~\cite{dai2017bundlefusion,10.1145/2487228.2487237}. 
As sensor technologies advance and computing resources scale, from commodity RGB-D cameras and large-scale LiDAR capture to real-time neural rendering systems, 3D perception is becoming increasingly practical and ubiquitous~\cite{zhang2012kinect,behleydataset,kerbl20233dgaussiansplattingrealtime}.

Unlike 2D vision, the field of 3D vision is fundamentally more complex, both in its data structures and in its learning pipelines~\cite{lahoud20223dvisiontransformerssurvey,lu2022transformers3dpointclouds,he2024deeplearningbased3d}. 
It spans a wide range of data representations, including point clouds, meshes, voxel grids, RGB-D images, multi-view images, CAD models, neural implicit fields, and 3D Gaussians, each with its own structural assumptions, learning pipelines, and computational trade-offs~\cite{qi2017pointnet,10.1145/2487228.2487237,wang2017ocnn,song2015sunrgbd,willis2020fusion,mescheder2019occupancy,kerbl20233dgaussiansplattingrealtime}. 
At the same time, downstream tasks range from reconstruction and segmentation to pose estimation and scene generation~\cite{dai2017bundlefusion,gupta2014learning,kendall2015posenet,dreamfusion,magic3d}. 
This diversity creates a steep learning curve for new researchers entering the domain~\cite{lahoud20223dvisiontransformerssurvey,gao2025nerfneuralradiancefield,chen2025survey3dgaussiansplatting}.

While there exist many task-specific papers and tutorials, most existing reviews remain architecture-centric, representation-centric, or task-specific, rather than offering a unified and data-centric view that connects data structures, benchmark datasets, and modeling paradigms in one framework~\cite{lahoud20223dvisiontransformerssurvey,lu2022transformers3dpointclouds,gao2025nerfneuralradiancefield,chen2025survey3dgaussiansplatting,li2024advances3dgenerationsurvey,Qian_2022}.

\textbf{In this cookbook}, we aim to bridge this gap by providing a unified, data-centric perspective on 3D vision. Our contributions are threefold:
\begin{itemize}
  \item We offer a high-level map of how 3D data are represented, stored, and processed in computers and machine learning systems, covering major formats such as point clouds, meshes, voxel grids, RGB-D images, CAD models, implicit fields, and 3D Gaussians within one unified view~\cite{qi2017pointnet,10.1145/2487228.2487237,wang2017ocnn,song2015sunrgbd,willis2020fusion,mescheder2019occupancy,kerbl20233dgaussiansplattingrealtime}.
 \item We highlight how datasets and benchmarks have not only enabled fair evaluation but also actively shaped the evolution of 3D learning paradigms by defining data structures, supervision formats, and scalability constraints~\cite{dai2017scannet,zheng2020structured3dlargephotorealisticdataset,willis2020fusion,yeshwanth2023scannethighfidelitydataset3d,grauman2024egoexo4d}.
  \item We situate emerging trends, such as 2D-supervised 3D learning, neural implicit fields, and the extension of 3D vision to 4D scene understanding and world modeling, within a broader narrative of efficiency, fidelity, and accessibility~\cite{dreamfusion,makeit3d,Latent-nerf,10.1145/3503250,kerbl20233dgaussiansplattingrealtime,wu20244dgaussiansplattingrealtime,bar2025navigationworldmodels}.
\end{itemize}

By distilling the field’s complexity into a structured map, we hope to make 3D vision more approachable, interpretable, and navigable for students and practitioners entering this rapidly expanding area~\cite{he2024deeplearningbased3d,li2024advances3dgenerationsurvey}.

\label{introduction}

\section{Scope of the Paper}

We specify the concrete scope and positioning of this survey. Our coverage spans three core axes:

\begin{itemize}
  \item \textbf{Data Representations:} We review the major data forms in 3D vision—point clouds, meshes, voxel grids, RGB-D and multi-view images, CAD/B-Rep models, neural implicit fields, and 3D Gaussian—and analyze their efficiency–fidelity trade-offs.
  \item \textbf{Datasets and Benchmarks:} We explor the dataset ecosystem across modalities and tasks, emphasizing how benchmark design both enables progress and constrains model development.
  \item \textbf{Modeling Paradigms:} We summarize classical geometry-based pipelines and modern neural approaches, including 2D-supervised 3D learning, implicit neural fields, and 4D video/world modeling.
\end{itemize}

Our review differs from existing reviews in both scope and perspective. Architecture-centric works ~\cite{lahoud20223dvisiontransformerssurvey, lu2022transformers3dpointclouds} focus on network families but not on the dataset–representation nexus. Topic-centric summaries ~\cite{gao2025nerfneuralradiancefield, chen2025survey3dgaussiansplatting} provide depth on one paradigm while leaving other representations disconnected. Task-oriented overviews ~\cite{s24072314,li2024advances3dgenerationsurvey, he2024deeplearningbased3d, Qian_2022}  offer detailed taxonomies for individual applications but seldom consider supervision strategies or cross-task scalability. Finally, mechanism-focused treatments~\cite{kato2020differentiablerenderingsurvey} analyze rendering pipelines in isolation, whereas in our cookbook differentiable rendering is treated only as one component of a broader spectrum.

\label{scope}

\section{A Taxonomy of 3D Representations}

\begin{figure*}[t]
    \centering
    \captionsetup[subfigure]{justification=centering}
    \begin{subfigure}[t]{0.24\textwidth}
        \centering
        \includegraphics[width=\linewidth]{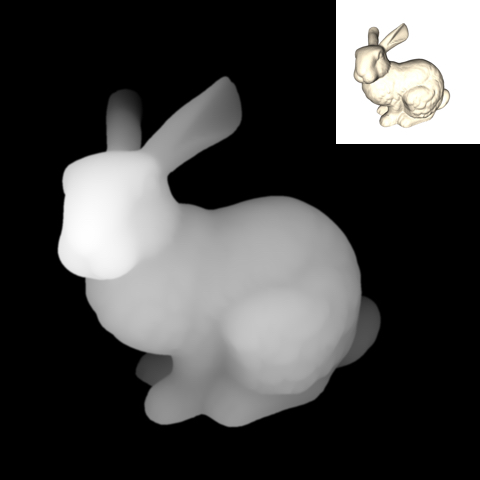}
        \caption{RGB-D}
        \label{fig:rep_rgbd}
    \end{subfigure}\hfill
    \begin{subfigure}[t]{0.24\textwidth}
        \centering
        \includegraphics[width=\linewidth]{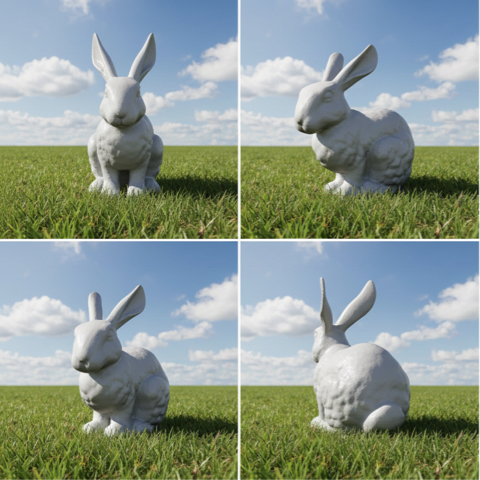}
        \caption{Multi-view Images}
        \label{fig:rep_multiview}
    \end{subfigure}\hfill
    \begin{subfigure}[t]{0.24\textwidth}
        \centering
        \includegraphics[width=\linewidth]{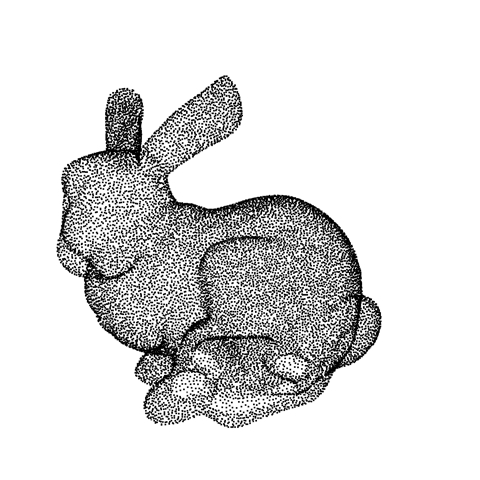}
        \caption{Point Cloud}
        \label{fig:rep_pointcloud}
    \end{subfigure}\hfill
    \begin{subfigure}[t]{0.24\textwidth}
        \centering
        \includegraphics[width=\linewidth]{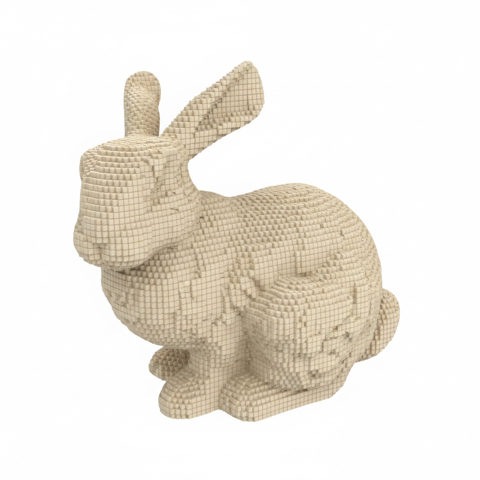}
        \caption{Voxels}
        \label{fig:rep_voxels}
    \end{subfigure}

    \par\medskip

    \begin{subfigure}[t]{0.24\textwidth}
        \centering
        \includegraphics[width=\linewidth]{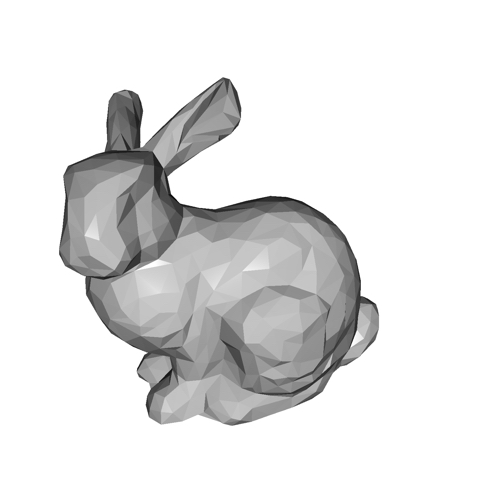}
        \caption{Mesh}
        \label{fig:rep_mesh}
    \end{subfigure}\hfill
    \begin{subfigure}[t]{0.24\textwidth}
        \centering
        \includegraphics[width=\linewidth]{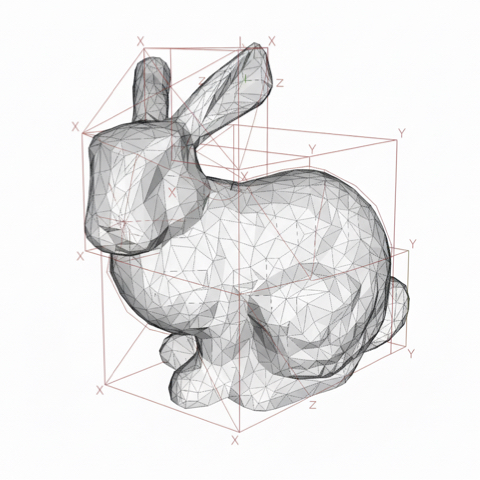}
        \caption{CAD}
        \label{fig:rep_cad}
    \end{subfigure}\hfill
    \begin{subfigure}[t]{0.24\textwidth}
        \centering
        \includegraphics[width=\linewidth]{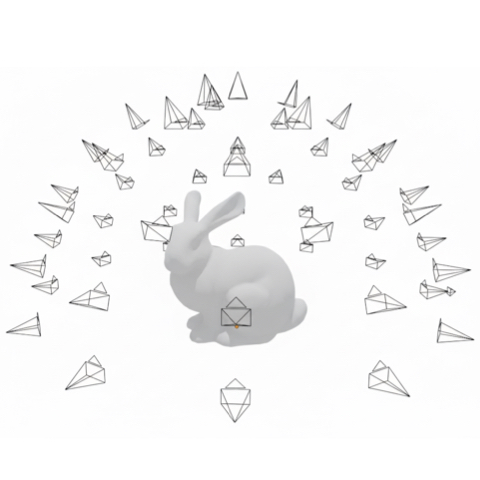}
        \caption{Implicit Field}
        \label{fig:rep_implicit}
    \end{subfigure}\hfill
    \begin{subfigure}[t]{0.24\textwidth}
        \centering
        \includegraphics[width=\linewidth]{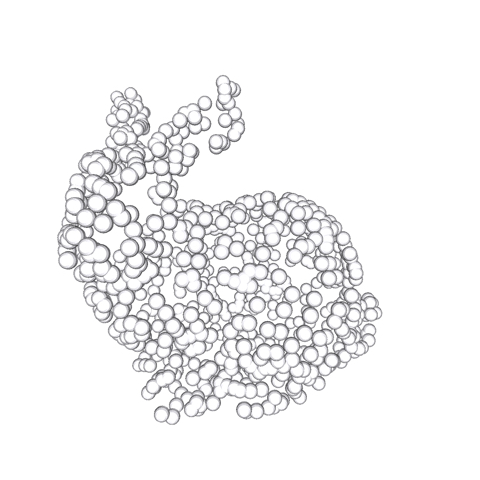}
        \caption{3D Gaussians}
        \label{fig:rep_gaussians}
    \end{subfigure}
    \caption{Various 3D representations of the Stanford bunny~\cite{10.1145/192161.192241}, including RGB-D, multi-view images, point cloud, voxels, mesh, CAD, implicit fields, and 3D Gaussians. These formats illustrate the diversity of 3D data modalities commonly used in benchmarks and learning frameworks.}
    \label{fig:placeholder}
\end{figure*}

\label{sec:landscape}

3D vision relies on diverse data representations—voxel grids, point clouds, implicit fields, and 3D Gaussians—each tailored to specific tasks like reconstruction and recognition. This section categorizes these representations by structure and efficiency and how each data type is acquired.

\subsection{RGB-D}

RGB-D data integrates RGB color images with per-pixel depth maps, capturing both appearance and geometry in a structured 2.5D format. For each pixel \((u, v)\) in the 2D image grid, the RGB value is denoted by \(\mathbf{c}(u, v) \in \mathbb{R}^3\) and the corresponding depth by \(d(u, v) \in \mathbb{R}\). The 3D point \(\mathbf{p} = (x, y, z)\) can be recovered via:
\[
\mathbf{p} = d(u, v) \cdot \mathbf{K}^{-1} \cdot [u, v, 1]^T
\]
where \(\mathbf{K}\) is the camera intrinsic matrix. This projection enables efficient 2D CNN processing of 3D data with a computational complexity of \(O(H \times W)\), where \(H \times W\) is the image resolution. 

RGB-D data is typically acquired using sensors such as Microsoft Kinect~\cite{zhang2012kinect,izadi2011kinectfusion}, Intel RealSense, or Structure Sensor. The depth map encodes the distance from the camera to visible surfaces in the scene, offering structured 3D geometry at the pixel level~\cite{silberman2012indoor,song2015sunrgbd,xiao2013sun3d,yeshwanth2023scannethighfidelitydataset3d}. Owing to its compactness and ease of use, RGB-D has become a widely adopted format in various 3D vision tasks, including indoor scene understanding~\cite{dai2017scannet,henry2012rgbdmapping,gupta2014learning,choi2015robust,cao2022monoscenemonocular3dsemantic}, pose estimation~\cite{shotton2011realtime,kendall2015posenet,song2016deep,teed2020deepv2d}, and SLAM~\cite{mur2017orbslam2,dai2017bundlefusion,runz2018cofusion,jin2024tensoirtensorialinverserendering}.

\subsection{Point Clouds}

A point cloud is a set of discrete points in 3D space, typically captured by LiDAR, RGB-D sensors, or photogrammetry~\cite{qi2018frustum}. It is defined as
\[
\{\mathbf{p}_i = (x_i, y_i, z_i) \in \mathbb{R}^3 \mid i = 1, \dots, N\}
\]
with optional attributes like color or normals. Processing complexity depends on the architecture: PointNet~\citep{qi2017pointnet} operates in \(O(N)\), while Transformer-based models like PointTransformer~\citep{zhao2021pointtransformer} scale as \(O(N^2)\). State-space models, such as PointMamba~\citep{liang2024pointmambasimplestatespace}, achieve \(O(N)\) complexity by leveraging structured state transitions.

The field began with PointNet/PointNet++~\cite{qi2017pointnet,qi2017pointnetpp}, which introduced point-wise and hierarchical feature extraction. Since then, a wide range of methods have been proposed for registration~\cite{Aoki_2019_CVPR,sarode2019pcrnetpointcloudregistration,9286491,Qin_2022_CVPR,9855233}, classification and segmentation~\cite{9880161,Zhao_2021_ICCV,Lai_2022_CVPR,Qiu_2022_ACCV,9861747,Guo_2021,Kolodiazhnyi_2024_CVPR} using deep learning or Transformer-based architectures.
Most recently, state-space models have emerged as efficient alternatives to Transformers. Oneformer3d~\cite{Kolodiazhnyi_2024_CVPR},
PointMamba~\cite{liang2024pointmambasimplestatespace}, Point Transformer~\cite{zhao2021pointtransformer,wu2022pointtransformerv2grouped,Wu_2024_CVPR} and other works~\cite{liu2024pointmambanovelpoint,zhang2024pointcloudmambapoint, wang2024pointrambahybridtransformermambaframework} significantly reduce computational cost while achieving competitive or superior performance, marking a new trend in point cloud modeling.

Point clouds can be acquired either directly or indirectly. Direct acquisition uses LiDAR or RGB-D sensors that measure range and back-project observations into 3D coordinates, yielding sparse outdoor scans or organized indoor point sets~\cite{behleydataset,qi2018frustum,izadi2011kinectfusion}. Indirect acquisition reconstructs 3D points from image collections via SfM and MVS/photogrammetry, and multi-view or multi-session captures are often merged through SLAM or global registration into a common coordinate frame~\cite{schoenberger2016sfm,schoenberger2016mvs,dai2017bundlefusion}. In synthetic benchmarks, point clouds are also frequently generated by sampling surfaces from meshes or CAD models, which provides clean geometry with controllable density and annotations~\cite{chang2015shapenetinformationrich3dmodel,willis2020fusion}.

\subsection{Voxels}
Voxel grids divide 3D space into uniform cells, each of which can store occupancy, color, density, or semantic information~\cite{wu2015shapenets,wang2017ocnn,cicek2016unet}. Their regular structure makes them naturally compatible with 3D convolutional neural networks~\cite{maturana2015voxnet,riegler2017octnet,graham2018sparseconvnet,choy2019minkowski}, and they are therefore widely used in volumetric reconstruction, segmentation, and object modeling~\cite{dai2018scancomplete,liu2020nsvf,yu2021plenoctreesrealtimerenderingneural,chen2022tensorf,sun2022directvoxelgridoptimization,jin2024tensoirtensorialinverserendering}.

Voxel grids discretize a 3D volume into a grid of size $N \times N \times N$, where each voxel at position $(x, y, z)$ is assigned a value $v(x, y, z)$. For binary occupancy, this is defined as:
\[
v(x, y, z) =
\begin{cases} 
1, & \text{if occupied} \\
0, & \text{otherwise}
\end{cases}
\]

For continuous attributes such as density or RGB color, the voxel value is given by $v(x, y, z) \in \mathbb{R}^k$, where $k$ denotes the dimensionality of the attribute vector.

Voxel data are rarely sensed directly. Instead, they are typically obtained either by \textit{voxelizing} meshes, CAD surfaces, or dense point clouds into occupancy or attribute grids, or by volumetric fusion of multi-view depth observations in TSDF/occupancy volumes from RGB-D or LiDAR scans aligned across viewpoints~\cite{chang2015shapenetinformationrich3dmodel,wu2015shapenets,izadi2011kinectfusion,dai2017bundlefusion,10.1007/978-3-319-11755-3_40,FISHER2021103755}. Synthetic benchmarks often produce voxels by rasterizing clean CAD assets, whereas real-scene datasets derive them from fused sensor measurements and then optionally attach color or semantic labels.


\subsection{Meshes}

Meshes provide a structured surface representation for modeling 3D geometry using vertices, edges, and faces. By explicitly encoding both shape and topology, meshes are well suited for applications such as graphics rendering, CAD design, and physical simulation~\cite{10.1145/2487228.2487237, kazhdan2006poisson,maturana2015voxnet,qi2017pointnetdeeplearningpoint}. 


Despite their expressiveness and compactness, the irregular structure of meshes makes them challenging to process using standard deep learning frameworks, which are generally optimized for grid-like data. As a result, many pipelines convert meshes to point clouds or voxels before learning~\cite{chang2015shapenetinformationrich3dmodel,wu2015shapenets,qi2017pointnet,maturana2015voxnet}. Direct mesh networks such as MeshCNN alleviate this mismatch, but remain more specialized than point- or voxel-based backbones~\cite{hanocka2019meshcnn}.

Meshes are commonly acquired in several ways. Active 3D scanners can capture multiple range images that are aligned and stitched into polygonal surfaces~\cite{10.1145/192161.192241}. RGB-D reconstruction systems instead fuse many depth frames into a volumetric field and then extract a surface mesh, as in KinectFusion, DynamicFusion, and BundleFusion~\cite{izadi2011kinectfusion,newcombe2015dynamicfusion,dai2017bundlefusion}. In photogrammetry pipelines, camera poses and dense geometry are recovered from RGB images via SfM/MVS, after which a mesh is reconstructed from the resulting point cloud using surface reconstruction methods such as Poisson reconstruction~\cite{schoenberger2016sfm,schoenberger2016mvs,kazhdan2006poisson,10.1145/2487228.2487237}. Many benchmark meshes are also obtained by tessellating CAD or artist-created assets into triangles before downstream learning~\cite{chang2015shapenetinformationrich3dmodel,willis2020fusion}.

\begin{table*}[t]
\centering
\caption{Summary of common 3D data representations.}
\setlength{\tabcolsep}{5pt}
\resizebox{\textwidth}{!}{%
\begin{tabular}{lcccl}
\toprule
\textbf{Representation} & \textbf{Structure} & \textbf{Efficiency} & \textbf{Fidelity} & \textbf{Applications} \\
\midrule
RGB-D             & 2.5D Grid (RGB + Depth)             & High       & Medium        & SLAM, indoor mapping, pose \\
Multi-view & 2D Views + Poses                    & High       & High$^*$      & SfM, MVS, NeRF input \\
Point Cloud       & Unstructured 3D Points              & High       & Low--Medium   & Detection, mapping, robotics \\
Mesh              & Vertex--Edge--Face Graph            & Medium     & High          & Modeling, animation, simulation \\
Voxel Grid        & Dense 3D Lattice                    & Low        & Medium        & Volumetric CNN, segmentation \\
Implicit Field    & Neural Function $f(x)$              & Low        & Very High     & View synthesis, scene modeling \\
3D Gaussians      & Sparse 3D Gaussian Distributions    & Very High  & High          & Real-time NeRF-style rendering \\
CAD Model         & Parametric Surfaces (NURBS)         & Very High  & Very High     & CAD design, reverse engineering \\
\bottomrule
\end{tabular}%
}

\vspace{4pt}
\raggedright
\footnotesize{$^*$~Fidelity refers to high visual fidelity (appearance); geometric structure must be inferred.}
\label{tab:3d-data-summary}
\end{table*}

\subsection{CAD}
\textbf{Computer-Aided Design (CAD)} models describe 3D shapes using \emph{smooth, mathematically defined surface patches}, most commonly through non-uniform rational B-splines (NURBS)~\cite{PieglTiller1997,Farin2002}. Each CAD model consists of a finite set of parametric patches:
\[
\mathcal{M}=\bigcup_{k=1}^{K} S_k,\quad
S_k:[0,1]^2 \to \mathbb{R}^3
\]
with each patch parameterized by~\cite{PieglTiller1997,deBoor1978,Cox1972}
\[
S(u,v)=
\frac{\displaystyle\sum_{i=0}^{n}\sum_{j=0}^{m} N_{i,p}(u)\,N_{j,q}(v)\,w_{ij}\,\mathbf{P}_{ij}}
     {\displaystyle\sum_{i=0}^{n}\sum_{j=0}^{m} N_{i,p}(u)\,N_{j,q}(v)\,w_{ij}},
\]
\[
(u,v)\in[0,1]^2
\]
where \(N_{i,p}, N_{j,q}\) are B-spline basis functions, \(\mathbf{P}_{ij}\) are control points, and \(w_{ij}>0\) are weights.~\cite{Cox1972,deBoor1978}  
This formulation enables closed-form evaluation of positions, derivatives, normals, and curvature, supporting high-fidelity rendering, exact intersections, and robust Boolean operations.~\cite{PieglTiller1997,PatrikalakisMaekawa2002,Hoffmann1989,Mantyla1988}

\textbf{Data acquisition:} CAD data are usually acquired through design workflows rather than direct sensing. In industrial practice, engineers create models interactively in CAD software, which naturally records sketches, constraints, feature histories, and final B-Rep/NURBS geometry; datasets such as Fusion 360 Gallery, SketchGraphs, and DeepCAD expose parts of this process for learning~\cite{willis2020fusion,seff2020sketchgraphslargescaledatasetmodeling,Wu2021DeepCAD}. Large research corpora are also assembled by harvesting existing repositories and converting STEP/B-Rep assets into canonical analytic patches or sequence-like representations, as in ABC and BRep2Seq~\cite{8954378,10.1093/jcde/qwae005}. When an editable model is needed for a real object, another route is scan-to-CAD retrieval and alignment, where images or reconstructed geometry are matched to a parametric template that can then be refined~\cite{Gumeli_2022}.

\subsection{Gaussians Splatting}

A 3D Gaussian is a continuous and compact primitive for representing spatial density, and has recently become a popular explicit representation for neural rendering~\cite{kerbl20233dgaussiansplattingrealtime}. Similar to point clouds, each Gaussian is defined by a position $\mu = (x, y, z)$ and a covariance matrix $\Sigma \in \mathbb{R}^{3 \times 3}$ that determines its shape and orientation in space. The probability density function is:

\[
f(\mathbf{x}) = \frac{1}{(2\pi)^{3/2} |\Sigma|^{1/2}} \exp\left(-\frac{1}{2} (\mathbf{x} - \mu)^T \Sigma^{-1} (\mathbf{x} - \mu)\right).
\]
To ensure $\Sigma$ is symmetric and positive semi-definite, it is decomposed as:
\[
\Sigma = R S S^T R^T,
\]
where $R$ is a rotation matrix and $S$ is a diagonal scaling matrix. In addition to geometry, each Gaussian carries:
\begin{itemize}
  \item \textbf{Opacity} $\alpha$, which controls how transparent the Gaussian appears.
  \item \textbf{Spherical Harmonics (SH)} coefficients, which model view-dependent color and enable realistic shading.
\end{itemize}
Each 3D Gaussian is typically initialized from an SfM point cloud~\cite{kerbl20233dgaussiansplattingrealtime,schoenberger2016sfm,Fu_2024_CVPR}, with position $\boldsymbol{\mu}_i$, unit covariance $\Sigma_i = I$, opacity $\alpha_i = 1$, and SH color $\mathbf{c}_i$ from the RGB value. During training, parameters are optimized via gradient descent to minimize the rendering loss $\mathcal{L}_{\text{render}}$:

\[
\theta_i^{(t+1)} = \theta_i^{(t)} - \eta \cdot \nabla_{\theta_i} \mathcal{L}_{\text{render}},
\]
where $\theta_i \in \{\boldsymbol{\mu}_i, \Sigma_i, \alpha_i, \mathbf{c}_i\}$.

Gaussian-splatting data are typically acquired from calibrated multi-view RGB images or videos rather than from a dedicated sensor. The standard pipeline first estimates camera poses and a sparse point cloud via SfM, optionally densifies geometry with MVS or depth priors, and then optimizes Gaussian positions, covariances, opacities, and colors against photometric rendering losses~\cite{kerbl20233dgaussiansplattingrealtime,schoenberger2016sfm,schoenberger2016mvs}. Recent methods reduce or remove the dependence on a full SfM/COLMAP-style initialization by learning pose-free or COLMAP-free Gaussian reconstruction from unposed image collections~\cite{Fu_2024_CVPR,ye2024poseproblemsurprisinglysimple,hong2025pf3platposefreefeedforward3d}. In online perception, Gaussians can also be updated incrementally from streaming observations, as demonstrated in Gaussian Splatting SLAM and dynamic 3DGS variants~\cite{matsuki2024gaussiansplattingslam,luiten2023dynamic3dgaussianstracking}.




\label{landscape}

\section{3D Learning Paradigms and Applications}

Modern 3D vision has increasingly shifted from explicit geometry pipelines toward learned systems that couple representation design, supervision, and practical utility~\cite{kato2020differentiablerenderingsurvey,wang2024dust3r,trellis}. To provide a clear conceptual map, this section is divided into two distinct parts. First, we discuss the core 3D learning and rendering paradigms that dictate how neural networks encode and supervise geometric data. Second, we explore how these fundamental paradigms are deployed across downstream applications, ranging from object reconstruction and scene generation to interactive 4D world models.

\subsection{Preliminary: Differentiable Rendering}

Early learning-based 3D methods often relied on direct 3D supervision, where losses such as Chamfer distance, Earth Mover's Distance, or volumetric TSDF errors were computed explicitly in 3D space~\cite{fan2016pointsetgenerationnetwork,qi2017pointnet,qi2017pointnetpp,wu2015shapenets,choy20163dr2n2unifiedapproachsingle}. Although conceptually simple, these objectives become computationally prohibitive for dense voxels or high-resolution surfaces. A pivotal transition came from differentiable rendering frameworks (e.g., Neural Mesh Renderer, Soft Rasterizer, OpenDR)~\cite{kato2017neural3dmeshrenderer,liu2019softrasterizerdifferentiablerenderer,10.1007/978-3-319-10584-0_11}. By backpropagating through the image formation process, these methods replace explicit 3D supervision with image-plane losses on color, depth, or silhouettes:
\[
\mathcal{L}_{\mathrm{photo}} = \sum_{i=1}^N \big\| I_i - \mathcal{R}(\mathcal{M}_\theta, P_i) \big\|^2
\]
where $\mathcal{R}$ is the differentiable rendering operator, $\mathcal{M}_\theta$ is the 3D representation, and $P_i$ denotes the camera parameters~\cite{kato2020differentiablerenderingsurvey}. The evolution of this rendering operator defines the computational limits of 3D learning:
\begin{itemize}
    \item \textbf{Volume Rendering (NeRFs):} Early continuous frameworks utilized ray-marching and volumetric integration. While physically principled, the dense multi-layer perceptron (MLP) queries along each ray made end-to-end training on high-resolution data computationally prohibitive~\cite{gao2025nerfneuralradiancefield}.
    \item \textbf{Tile-based Rasterization (3DGS):} The introduction of 3D Gaussian Splatting revolutionized the rendering bridge. By replacing implicit MLPs with explicit 3D Gaussians and utilizing a highly optimized, differentiable $\alpha$-blending rasterizer, 3DGS reduced rendering times from seconds to milliseconds. This breakthrough directly enabled the training of massive, feed-forward 3D foundation models~\cite{kerbl20233dgaussiansplattingrealtime}.
\end{itemize}

\subsection{Learning Paradigm for End-to-End Geometric Foundation Models:}

Building on image-plane supervision, image-aligned representations have emerged as a leading paradigm because they preserve dense per-pixel structure while keeping learning in the 2D domain~\cite{wang2024dust3r,wang2025vggtvisualgeometrygrounded,depthanything3}. Several foundational formulations define this space:
\begin{itemize}
    \item \textbf{DUSt3R~\cite{wang2024dust3r}:} Learns through confidence-weighted regression on image-aligned 3D outputs without explicit multi-view optimization at training time:
    \begin{equation}
        \mathcal{L}_{\mathrm{pmap}} = \sum_i \left( \| C_i \odot (P_i - P^*_i) \| - \alpha \log C_i \right)
    \end{equation}
    where $P_i$ and $P^*_i$ are predicted and ground-truth 3D points, and $C_i$ models aleatoric uncertainty.
    
    \item \textbf{VGGT~\cite{wang2025vggtvisualgeometrygrounded}:} Scales the image-aligned paradigm to large multi-view sets by jointly optimizing a multi-task objective for reusable geometric backbones:
    \begin{equation}
        \mathcal{L}_{\mathrm{total}} = \mathcal{L}_{\mathrm{camera}} + \mathcal{L}_{\mathrm{depth}} + \mathcal{L}_{\mathrm{pmap}} + \lambda \mathcal{L}_{\mathrm{track}}
    \end{equation}
    
    \item \textbf{RayZer~\cite{jiang2025rayzerselfsupervisedlargeview}:} Factorizes input into camera and scene representations to train entirely through 2D self-supervised reconstruction, without explicit 3D geometry:
    \begin{equation}
        \mathcal{L}_{\mathrm{RayZer}} = \| \hat{I}_{\mathrm{target}}(\hat{P}_{\mathrm{target}}) - I_{\mathrm{target}} \|_2^2
    \end{equation}
    
    \item \textbf{$\pi^3$~\cite{wang2025pi3scalablepermutationequivariantvisual}:} Enforces permutation-equivariant supervision over unordered image sets by optimizing local point maps ($X_i$) and relative poses ($T_{i \to j}$):
    \begin{equation}
        \mathcal{L}_{\pi^3} = \mathcal{L}_{\mathrm{local}}(X_i, X^*_i) + \mathcal{L}_{\mathrm{relative}}(T_{i \to j}, T^*_{i \to j})
    \end{equation}
    
    \item \textbf{Depth Anything 3~\cite{depthanything3}:} Collapses multiple geometric heads into a unified depth-plus-ray representation $R \in \mathbb{R}^{H \times W \times 6}$ (origin and direction):
    \begin{equation}
        \mathcal{L}_{\mathrm{DA3}} = \mathcal{L}_{\mathrm{depth}}(D, D^*) + \mathcal{L}_{\mathrm{ray}}(R, R^*)
    \end{equation}
\end{itemize}

\paragraph{Optimization via Generative Priors and Structured Latents:}

When explicit 3D data is scarce, learning paradigms shift toward distilling priors from large-scale 2D models or utilizing structured latent spaces. Methods like DreamFusion and Magic3D optimize neural fields through Score Distillation Sampling (SDS)~\cite{dreamfusion,magic3d}. More recently, models have moved toward \textbf{Native 3D Geometric Foundation Models}. TRELLIS learns structured 3D latents decodable into radiance fields, Gaussians, or meshes~\cite{trellis}. Concurrently, SAM 3D formulates learning as \textbf{Rectified Conditional Flow Matching (RCFM)}, uniquely breaking the 3D data barrier through a \textbf{Model-in-the-Loop (MITL)} data engine where generative outputs are human-vetted to create recursive supervision~\cite{chen2025sam3d}.

\paragraph{The Synergy of Reconstruction and Generation:}
Historically treated as separate domains, Geometric Foundation model now heavily couple reconstruction and generation. \textit{Generation for Reconstruction} utilizes generative priors (e.g., RCFM or diffusion) to hallucinate missing geometry in ill-posed, sparse-view settings~\cite{chen2025sam3d,shi2024mvdreammultiviewdiffusion3d}. Conversely, \textit{Reconstruction for Generation} extracts rigid geometric scaffolding to constrain generative models to physically consistent layouts. This synergy increasingly operates within shared latent spaces, enabling a continuous \textit{data flywheel} where synthetic generation and automated reconstruction mutually improve the training corpus~\cite{wang2024dust3r,chen2025sam3d,Jiang_2025_CVPR}.

\subsection{Downstream Applications}

The 3D vision field has also rapidly expanded its applicative scope by leveraging the rendering techniques, image-aligned representations, and End-to-End 3D Geometric Foundation Model.

\paragraph{3D Reconstruction:}
3D reconstruction seeks to recover object or scene geometry from visual inputs. Classical pipelines relied on Structure-from-Motion (SfM) and multi-view stereo~\cite{schoenberger2016sfm,Furu:2010:PMVS}, which are mathematically principled but brittle under sparse views or weak texture. Modern applications replace these bottlenecks entirely with the aforementioned image-aligned neural backbones, enabling robust, end-to-end recovery of point maps, depth, and cameras directly from uncalibrated imagery, even in zero-shot or single-view scenarios~\cite{wang2024dust3r,depthanything3,wang2025moge2accuratemonoculargeometry}.

\paragraph{3D Asset and Scene-Level Generation:}

To circumvent the slow per-prompt optimization of SDS, modern asset generation employs feed-forward multi-view reconstruction. Multi-view diffusion models synthesize view-consistent images, which Large Reconstruction Models (LRMs) instantly map into meshes, tri-planes, or Gaussians~\cite{shi2024mvdreammultiviewdiffusion3d,zero123,lrm,instantmesh,Hunyuan3D2.5}. Beyond isolated objects, applications are scaling to composition and layout. Frameworks like 3D-SceneDreamer and AnyHome target open-vocabulary generation of structured, navigable indoor environments with explicit room and object-level organization~\cite{Zhang_2024_CVPR,fu2024anyhomeopenvocabularygenerationstructured}.

\paragraph{3D Consistent Video Generation:}
Large video diffusion models (VDMs) generate visually stunning content but struggle to preserve stable geometry across time and camera motion. Applications in this domain focus on injecting 3D paradigms to regulate generation~\cite{hong2022cogvideo,wan2025}. \textit{3D Geometric Preference Alignment} uses 3D consistency as a reward signal, applying Direct Preference Optimization (DPO) based on epipolar Sampson distance or distilled geometric priors from 3D Geometric Foundation Model suppresses physically implausible in videos~\cite{kupyn2025epipolargeometryimprovesvideo,du2026videogpadistillinggeometrypriors}. \textit{Feature-Level Forcing} aligns latent diffusion features with depth or epipolar lines during denoising~\cite{wu2025geometryforcingmarryingvideo,tseng2023consistentviewsynthesisposeguided}. Furthermore, \textit{3D-Aware Control} conditions video synthesis on dense 3D trajectories (e.g., Diffusion as Shader), providing precise spatial manipulation over the generated motion~\cite{gu2025diffusionshader3dawarevideo}.

\paragraph{4D Rendering and 3D World Models:}

The application of 3D vision is expanding toward temporally persistent simulation. \textbf{4D Rendering} extends static Gaussian splatting with deformation fields, representing motion as structured 3D evolution rather than a sequence of 2D frames, enabling real-time rendering of dynamic topologies~\cite{wu20244dgaussiansplattingrealtime,wu2024cat4dcreate4dmultiview}. Extending this concept, \textbf{3D World Models} aim to predict future states for planning. Unlike 2D sequence rollouts, models like PointWorld and ParticleFormer push the state space into persistent 3D points or particles~\cite{huang2026pointworldscaling3dworld,huang2025particleformer3dpointcloud}. This ensures temporal consistency, strict multi-view faithfulness, and realistic physical interactions as evaluated by benchmarks like WorldSimBench~\cite{qin2024worldsimbenchvideogenerationmodels}.

\paragraph{Spatial Intelligence in Vision-Language-Action:}

The ultimate practical application of 3D world models lies in Embodied AI. Instead of mapping 2D image tokens directly to embodiment-specific motor outputs (e.g., joint torques), modern 3D-VLA systems ground perception, language, and robotic control in shared 3D representations~\cite{10.5555/3692070.3692890,hong20233dllm,xu2023pointllm}. By representing intent as 3D point flows or spatial trajectories, these frameworks dramatically improve viewpoint robustness, enable cross-embodiment generalization, and unlock complex spatial reasoning for physical agents~\cite{huang2026pointworldscaling3dworld}.
\label{application}

\section{Dataset and Benchmark}
\label{sec: challenges}

\definecolor{benchblue}{RGB}{70,107,176}
\definecolor{benchteal}{RGB}{68,162,153}
\definecolor{benchorange}{RGB}{224,159,62}

\pgfplotsset{
benchmarkbase/.style={
    tick label style={font=\scriptsize, color=gray!80!black},
    label style={font=\scriptsize, color=black!80},
    title style={font=\small, color=black!90},
    axis line style={draw=gray!40, thin},
    tick style={draw=gray!40, thin},
    nodes near coords,
    every node near coord/.append style={font=\tiny, color=black!70, /pgf/number format/fixed},
    every axis plot/.append style={draw=none, fill opacity=0.9},
    axis x line*=bottom,
    axis y line*=left,
    clip=false,
},
benchmarkybar/.style={
    benchmarkbase,
    ybar,
    ymin=0,
    ymajorgrids,
    grid style={gray!12, thin},
    enlarge x limits=0.06,
},
benchmarkxbar/.style={
    benchmarkbase,
    xbar,
    xmin=0,
    xmajorgrids,
    grid style={gray!12, thin},
    enlarge y limits=0.18,
}
}

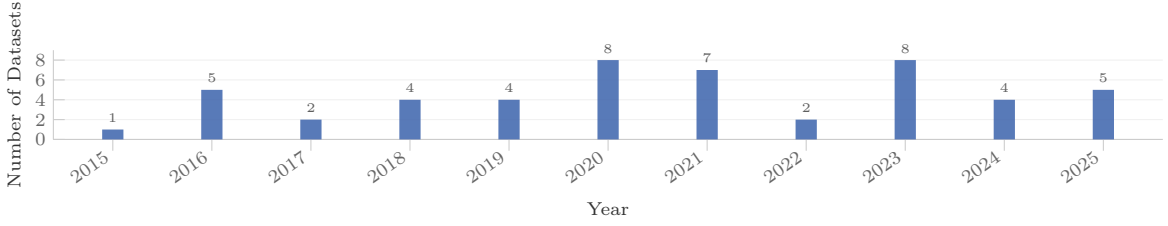
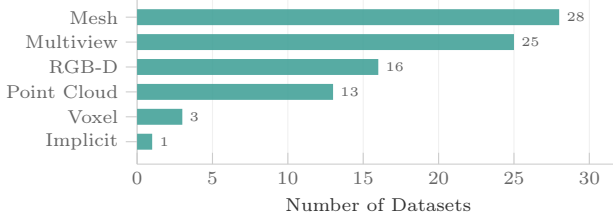
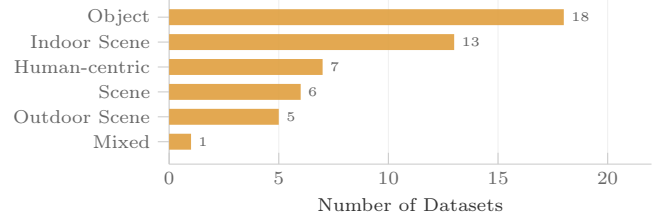
\begin{figure*}[t]
\centering
\scriptsize
\captionsetup{skip=2pt}
\captionsetup[subfigure]{singlelinecheck=false}
\begin{subfigure}[t]{0.98\textwidth}
\centering
\begin{tikzpicture}
\begin{axis}[
    benchmarkybar,
    width=\linewidth,
    height=0.17\textwidth,
    bar width=8pt,
    ylabel={Number of Datasets},
    xlabel={Year},
    ymax=9,
    symbolic x coords={2015,2016,2017,2018,2019,2020,2021,2022,2023,2024,2025},
    xtick=data,
    x tick label style={rotate=35, anchor=east},
]
\addplot[fill=benchblue] coordinates {
    (2015,1) (2016,5) (2017,2) (2018,4) (2019,4) (2020,8)
    (2021,7) (2022,2) (2023,8) (2024,4) (2025,5)
};
\end{axis}
\end{tikzpicture}
\caption{Number of datasets released each year, not exhaustive}
\label{fig:benchmark_year_chart}
\end{subfigure}
\par\vspace{0.4em}
\begin{subfigure}[t]{0.48\textwidth}
\centering
\begin{tikzpicture}
\begin{axis}[
    benchmarkxbar,
    width=\linewidth,
    height=0.48\linewidth,
    bar width=6pt,
    xlabel={Number of Datasets},
    xmax=32,
    symbolic y coords={Mesh,Multiview,RGB-D,Point Cloud,Voxel,Implicit},
    ytick=data,
    y dir=reverse,
    nodes near coords align={horizontal},
    point meta=x,
]
\addplot[fill=benchteal] coordinates {
    (28,Mesh)
    (25,Multiview)
    (16,RGB-D)
    (13,Point Cloud)
    (3,Voxel)
    (1,Implicit)
};
\end{axis}
\end{tikzpicture}
\caption{Dataset counts per modality. Modalities can overlap in one dataset, so the bars are not mutually exclusive.}
\label{fig:benchmark_modality_chart}
\end{subfigure}
\hfill
\begin{subfigure}[t]{0.48\textwidth}
\centering
\begin{tikzpicture}
\begin{axis}[
    benchmarkxbar,
    width=\linewidth,
    height=0.48\linewidth,
    bar width=6pt,
    xlabel={Number of Datasets},
    xmax=22,
    symbolic y coords={Object,Indoor Scene,Human-centric,Scene,Outdoor Scene,Mixed},
    ytick=data,
    y dir=reverse,
    nodes near coords align={horizontal},
    point meta=x,
]
\addplot[fill=benchorange] coordinates {
    (18,Object)
    (13,Indoor Scene)
    (7,Human-centric)
    (6,Scene)
    (5,Outdoor Scene)
    (1,Mixed)
};
\end{axis}
\end{tikzpicture}
\caption{Dataset counts by spatial granularity. Granularity can overlap in one dataset, so the bars are not mutually exclusive.}
\label{fig:benchmark_granularity_chart}
\end{subfigure}
\caption{Summary statistics for the 50 representative datasets listed in Tables~\ref{tab:datasets}. The release timeline shows in (a). The modality chart in (b) replaces the previous pie chart because benchmark modalities are multi-label rather than mutually exclusive. The granularity chart in (c) shows that object-centric and indoor-scene benchmarks currently dominate the landscape.}
\label{fig:combined_figure}
\end{figure*}

While \cref{sec:landscape} examined the structural spectrum of 3D representations, their practical impact is ultimately mediated through benchmark datasets, which establish learning objectives, task formulations, and evaluation protocols. We categorize existing datasets along four orthogonal axes: (1) \textbf{Data modality} (RGB-D, point cloud, mesh, multi-view images, implicit fields, Gaussians); (2) \textbf{Spatial granularity} (object-level, scene-level (indoor/outdoor), human-centric (face/hand/body), or mixed); (3) \textbf{Task formulation} (segmentation, correspondence, reconstruction, generation); and (4) \textbf{Temporal dimension} (static 3D versus dynamic 4D). This lens is increasingly important because recent benchmarks no longer merely collect data; they also encode the assumptions of modern 3D pipelines, from image-aligned reconstruction to 3DGS-native learning~\cite{yeshwanth2023scannethighfidelitydataset3d,ling2023dl3dv10klargescalescenedataset,Jiang_2025_CVPR,InteriorGS2025}.

\begingroup
\par\smallskip
\small
\setlength{\tabcolsep}{6pt}
\renewcommand{\arraystretch}{0.95}
\setlength{\LTpre}{0pt}
\setlength{\LTpost}{0pt}
\setlength{\LTcapwidth}{\textwidth}
\begin{xltabular}{\textwidth}{@{}l c >{\raggedright\arraybackslash\hspace{0pt}}X@{}}
\caption{Representative 3D datasets and benchmarks reviewed in this survey.}\label{tab:datasets}\\
\toprule
\textbf{Dataset} & \textbf{Year} & \textbf{Description} \\
\midrule
\endfirsthead
\multicolumn{3}{@{}l}{\small\itshape Table~\thetable\ (continued)}\\
\toprule
\textbf{Dataset} & \textbf{Year} & \textbf{Description} \\
\midrule
\endhead
\midrule
\multicolumn{3}{r@{}}{\footnotesize\itshape continued on next page}\\
\endfoot
\bottomrule
\endlastfoot
SAM 3D Body~\cite{yang2025sam3dbody} & 2025 & Promptable foundation model for full-body HMR \\
GigaHands~\cite{fu2025gigahands} & 2025 & 3D bimanual hand dataset with mesh and text labels \\
InteriorGS~\cite{InteriorGS2025} & 2025 & Synthetic indoor scenes with trajectories and labels \\
HPSketch~\cite{Fan2025HPSketch} & 2025 & History-based parametric CAD sketch dataset \\
CBF~\cite{10.1145/3731715.3733283} & 2025 & B-rep CAD models with base plate and three features \\
EgoExo4D~\cite{grauman2024egoexo4d} & 2024 & Egocentric/exocentric video dataset with 3D human pose \\
Parametric 20000~\cite{Parametric20000_2024} & 2024 & Multi-modal CAD shapes: point cloud, mesh, and B-Rep \\
WildRGB-D~\cite{xia2024rgbd} & 2024 & Real RGB-D object videos with 360° views and masks \\
BRep2Seq~\cite{10.1093/jcde/qwae005} & 2024 & B-rep solids paired with construction sequences \\
EgoHumans~\cite{khirodkar2023egohumansegocentric3dmultihuman} & 2023 & Multi-view egocentric 3D human--human interaction \\
Aria Synthetic Environments~\cite{pan2023ariadigitaltwinnew} & 2023 & Synthetic indoor scenes with device paths and labels \\
DL3DV-10K~\cite{ling2023dl3dv10klargescalescenedataset} & 2023 & Multi-view dataset over 65 scene types for view synthesis \\
PointOdyssey~\cite{zheng2023pointodysseylargescalesyntheticdataset} & 2023 & Synthetic videos for long-term point tracking \\
Aria Digital Twin~\cite{pan2023ariadigitaltwinnew} & 2023 & Egocentric dataset with 3D object \& human pose \\
ScanNet++~\cite{yeshwanth2023scannethighfidelitydataset3d} & 2023 & High-fidelity indoor scans with RGB-D and dense labels \\
Objaverse~\cite{objaverseXL} & 2023 & Large 3D mesh--text pairs for multimodal learning \\
DIVA-360~\cite{Lu_2024} & 2023 & Multi-view dataset for dynamic neural fields \\
H3WB~\cite{Zhu_2023_ICCV} & 2022 & Whole-body 3D keypoints for Human3.6M \\
Kubric~\cite{greff2021kubric} & 2022 & Synthetic generator for scenes/objects with annotations \\
Amazon Berkeley Objects~\cite{collins2022abo} & 2021 & Real-world objects with CAD, materials, and images \\
HM3D~\cite{ramakrishnan2021habitatmatterport3ddatasethm3d} & 2021 & Building-scale indoor meshes with high fidelity \\
Fusion 360 Gallery Dataset~\cite{willis2020fusion} & 2021 & CAD dataset with meshes and assembly data \\
CO3Dv2~\cite{reizenstein21co3d} & 2021 & Multi-view images + point clouds, 50 object categories \\
HyperSim~\cite{roberts2021hypersimphotorealisticsyntheticdataset} & 2021 & Photorealistic indoor scenes with dense annotations \\
Habitat 2.0~\cite{szot2022habitat20traininghome} & 2021 & Interactive apartments with articulated objects \\
StrobeNet~\cite{zhang2021strobenetcategorylevelmultiviewreconstruction} & 2021 & Articulated objects with joints and implicit shapes \\
RELLIS-3D~\cite{jiang2020rellis3d} & 2020 & Multi-sensor dataset for outdoor segmentation \\
Virtual KITTI 2~\cite{cabon2020virtualkitti2} & 2020 & Synthetic KITTI clones with varied conditions \\
FaceScape~\cite{yang2020facescape} & 2020 & High-quality textured 3D face scans with expressions \\
3D-FRONT~\cite{fu20213dfront3dfurnishedrooms} & 2020 & Synthetic furnished rooms with semantic layouts \\
3D-FUTURE~\cite{fu20203dfuture3dfurnitureshape} & 2020 & CAD furniture models with aligned textures \\
SketchGraphs~\cite{seff2020sketchgraphslargescaledatasetmodeling} & 2020 & CAD sketches as geometric-constraint graphs \\
Structured3D~\cite{zheng2020structured3dlargephotorealisticdataset} & 2020 & Synthetic photorealistic scenes with structure labels \\
Mapillaryc~\cite{ertler2020mapillarytrafficsigndataset} & 2020 & Street-level dataset for place recognition \\
ScanObjectNN~\cite{uy2019revisitingpointcloudclassification} & 2019 & Real-world point clouds with clutter and occlusion \\
ABC~\cite{8954378} & 2019 & CAD models with analytic geometry and labels \\
BlendedMVS~\cite{yao2020blendedmvslargescaledatasetgeneralized} & 2019 & MVS dataset mixing rendered and real images \\
Replica~\cite{replica19arxiv} & 2019 & Realistic indoor reconstructions with dense labels \\
3DPW~\cite{vonMarcard20183dpw} & 2018 & In-the-wild video + 3D ground truth from IMUs \\
RealEstate10K~\cite{zhou2018stereomagnificationlearningview} & 2018 & YouTube real-estate videos with camera poses \\
MegaDepth~\cite{MegaDepthLi18} & 2018 & Internet photos with dense depth from SfM/MVS \\
DeepMVS~\cite{DeepMVS} & 2018 & Synthetic MVS images with ground-truth matching \\
ScanNet~\cite{dai2017scannet} & 2017 & RGB-D scans with semantic meshes and CAD alignment \\
Matterport3D~\cite{chang2017matterport3dlearningrgbddata} & 2017 & RGB-D scans with panoramic views and segmentation \\
Thingi10K~\cite{Thingi10K} & 2016 & 3D printable meshes for shape analysis \\
Semantic3D~\cite{hackel2017isprs} & 2016 & Outdoor point clouds (\textasciitilde 4B pts) with labels \\
SceneNN~\cite{scenenn-3dv16} & 2016 & Indoor RGB-D reconstructions with semantic labels \\
Object Scans~\cite{Choi2016} & 2016 & Real object scans from diverse environments \\
Virtual KITTI~\cite{gaidon2016virtualworldsproxymultiobject} & 2016 & Synthetic KITTI sequences with full labels \\
ShapeNet~\cite{chang2015shapenetinformationrich3dmodel} & 2015 & Large CAD dataset with rich annotations \\
\end{xltabular}
\par\vspace{2pt}
{\footnotesize This list is not exhaustive; we will maintain an updated version on our GitHub.\par}
\par\smallskip
\endgroup

As illustrated in Figure~\ref{fig:combined_figure}, dataset releases have surged over the past decade, reflecting both advances in sensor technology and growing demand for 3D benchmarks. The updated counts from the show two especially active release since 2020, suggesting that benchmark growth is driven not by a steady linear trend but by bursts tied to new sensing pipelines and model families. Recent examples already show three distinct scaling directions: high-fidelity real capture in curated settings (e.g., ScanNet++~\cite{yeshwanth2023scannethighfidelitydataset3d}), in-the-wild object-centric RGB-D acquisition (e.g., WildRGB-D~\cite{xia2024rgbd}), and large synthetic or semi-synthetic corpora for long-range correspondences and scene reconstruction, such as PointOdyssey~\cite{zheng2023pointodysseylargescalesyntheticdataset} and DL3DV-10K~\cite{ling2023dl3dv10klargescalescenedataset}. Modality coverage also remains highly uneven: mesh-backed datasets (28/50) and multi-view benchmarks (25/50) are much more common than voxel (3/50) or implicit-field (1/50) datasets. Spatially, object-centric (18) and indoor-scene (13) datasets dominate, while mixed and outdoor settings remain comparatively scarce. We provide a comprehensive breakdown of these statistics in Table~\ref{tab:datasets}, further underscores this fragmentation.

Another recent shift is that benchmark construction itself is becoming model-aware. MegaSynth uses synthesized scenes to scale pretraining for scene reconstruction, while InteriorGS provides semantically labeled indoor scenes directly in the 3D Gaussian Splatting regime rather than only in meshes or point clouds~\cite{Jiang_2025_CVPR,InteriorGS2025}. At the evaluation level, suites such as WorldSimBench suggest that future 3D/4D benchmarks must assess not only reconstruction fidelity but also whether generative models behave like usable simulators under long-horizon, physically grounded tasks~\cite{qin2024worldsimbenchvideogenerationmodels}.

Despite rapid progress, these trends expose fundamental gaps. Current benchmarks still lack large-scale, multi-modal coverage that simultaneously supports heterogeneous representations (e.g., points, meshes, splats, and images), temporal consistency, and open-world generalization. Scene datasets such as ScanNet++~\cite{yeshwanth2023scannethighfidelitydataset3d} and DL3DV-10K~\cite{ling2023dl3dv10klargescalescenedataset} emphasize geometry and view diversity, object datasets such as WildRGB-D~\cite{xia2024rgbd} emphasize real-world capture, and synthetic datasets such as PointOdyssey~\cite{zheng2023pointodysseylargescalesyntheticdataset}, MegaSynth~\cite{Jiang_2025_CVPR}, and InteriorGS~\cite{InteriorGS2025} emphasize controllable scale or representation alignment; few benchmarks combine all of these attributes within one unified protocol. Bridging these gaps will require datasets that balance scale with diversity, minimize annotation overhead, and support both synthetic and in-the-wild scenarios—providing the foundation for robust and generalizable 3D/4D learning.

\label{challenges}

\section{Conclusion}

We offer a data-centric view of 3D vision, unifying \emph{representations, datasets, and learning paradigms} into a coherent framework. By tracing the trade-offs among different data representations, we clarify how efficiency, fidelity, and scalability jointly shape representation design. We further mapped the benchmark landscape and reviewed the evolution from geometry-based methods to neural implicit fields and 2D-supervised pipelines, highlighting how supervision regimes co-evolve with data availability.  

Despite the progress, key challenges remain: fragmented datasets hinder fair comparison, voxel- and mesh-based approaches struggle with scalability, and generalization beyond curated domains is still limited. At the same time, emerging areas—such as 4D spatiotemporal reasoning, physics-aware modeling, and world-consistent video generation—call for tighter integration of 3D priors with multimodal and physical signals.  

Looking ahead, we see three promising directions: (i) unified benchmarks and evaluation protocols that span objects, scenes, and dynamics; (ii) cross-modal and 2D-supervised learning strategies that exploit large-scale image data while preserving geometric grounding; and (iii) scalable, real-time representations, from Gaussian splats to parametric CAD, that balance efficiency with fidelity.  

\label{limitation}

\clearpage
\newpage
\bibliographystyle{ieeenat_fullname}
\bibliography{main}

\end{document}